\documentclass[letterpaper]{article} 
\usepackage{aaai2026}  
\usepackage{times}  
\usepackage{helvet}  
\usepackage{courier}  
\usepackage[hyphens]{url}  
\usepackage{graphicx} 
\usepackage{natbib}  
\usepackage{caption} 
\usepackage{algorithm}
\usepackage{algorithmic}

\usepackage{multirow}  
\usepackage{pifont}
\usepackage{subcaption}

\usepackage{newfloat}
\usepackage{listings}
\DeclareCaptionStyle{ruled}{labelfont=normalfont,labelsep=colon,strut=off} 
\floatstyle{ruled}
\newfloat{listing}{tb}{lst}{}
\floatname{listing}{Listing}
\title{Beyond Manually Designed Pruning Policies with Second-Level Performance Prediction: A Pruning Framework for LLMs}
\author{
    Zuxin Ma\textsuperscript{1,2},
    Yunhe Cui\textsuperscript{1,2},
    Yongbin Qin\textsuperscript{1,2}\thanks{{Corresponding author}}
}
\affiliations{
    \textsuperscript{\rm 1}State Key Laboratory of Public Big Data, Guizhou University, China\\
    \textsuperscript{\rm 2}School of Computer Science and Technology, Guizhou University, China\\
    \{gs.zxma24, yhcui, ybqin\}@gzu.edu.cn
}

\usepackage{bibentry}

\begin{document}

\maketitle

\begin{abstract}
Non-uniform structured network pruning methods can effectively reduce Large Language Model (LLM) size by eliminating redundant channels or layers, offering lower performance degradation than uniform strategies. However, existing non-uniform methods rely heavily on manually designed pruning policies (e.g., layer importance and scaling factors), and therefore cannot efficiently adapt to scenarios with dynamic pruning ratio requirements. Additionly, a critical bottleneck\textemdash the time-consuming evaluation of pruning policies\textemdash further limits the feasibility of iteratively and dynamically finding optimal pruning policies. To address these limitations, we propose \textbf{PPF} (\textbf{P}redictive \textbf{P}runing \textbf{F}ramework), a novel pruning framework for LLMs that eliminates manual design dependencies via second-level performance prediction. PPF not only supports real-time pruning decisions under dynamic pruning ratios but is also applicable to static pruning scenarios. It employs an agent for producing adaptive and real-time pruning actions, while a lightweight performance predictor that can evaluate a pruning policy in seconds, significantly speeding up the iterative optimization process. Experiments on Llama2-7B and Llama3-8B show that PPF can generate dynamic/static pruning policies and it reduces perplexity by up to 33.4\% (dynamic pruning) and 84.78\% (static pruning) over existing methods, outperforming manually designed pruning policies. The performance predictor achieves second-level performance prediction with high accuracy (prediction error $<$ 0.0011). It reduces the mean evaluation latency from minute-level (1 minute and 38.02 seconds of test-set evaluation methods) to second-level (1.52 seconds), achieving over 64$\times$ speedup. Our code will be available at https://github.com/Ma-zx/PPF .
\end{abstract}


%
%
%
\section{Introduction}
\subsubsection{Background.}
Large Language Models (LLMs) exhibit remarkable abilities in understanding and generating text. However, their large parameter size creates significant redundancy. In edge environments, deploying production-level LLMs faces challenges due to their high requirement on computational resources and memory capacity. To address this challenge, structured pruning technique \cite{ma23llmpruner} has been proposed to significantly decrease both model size and computational cost \cite{MichelLN19Are}, making LLMs more suitable for edge deployment. Meanwhile, non-uniform structured pruning \cite{yin2024outlier, lu24alpha} typically causes less performance degradation than uniform structured pruning \cite{ma23llmpruner} in LLMs.
\subsubsection{Motivation.}
However, previous non-uniform pruning methods relied on manual design or expert knowledge to explore numerous pruning policy combinations \cite{yin2024outlier, lu24alpha}. This typically requires manually determining how to calculate the layer importance and scaling factors between importance scores and pruning ratios. When pruning ratios frequently change due to user or environmental factors \cite{Jang25Edge, EcclesRKSV24DNNShifter}, existing methods struggle to quickly provide optimal pruning policies. Therefore, the above gap motivates us to explore an agent-based pruning framework that can automatically search for optimal pruning policies under varying pruning ratios. However, this introduces a new issue: evaluating a single pruning policy is time-consuming, which becomes a fundamental bottleneck preventing the feasible iterative and dynamic optimization of pruning policies.
\subsubsection{Challenges.}
Based on this motivation, we need a pruning framework that can respond to dynamic pruning ratio changes in real-time and support fine-grained static pruning optimization. A key challenge lies in two requirements: developing the agents to learn generalizable pruning policies from limited samples across all dynamic ranges, and enabling it escape local optima for finer searches at static ratios. 
Additionally, another critical challenge is that our proposed framework necessitates an evaluator, which has to offline characterize relationships between pruning policies and model performance, while predicting the performances of candidate pruning policies in real-time.
\subsubsection{Contribution.}
To address these challenges, we proposed \textbf{PPF}. Our contributions are as follows:
\begin{itemize}
\item We propose PPF, a novel pruning framework that eliminates manual design dependencies by second-level performance prediction. This capability enables real-time decision making under varying pruning ratios and fine-grained search under fixed pruning ratios. PPF is the first structured pruning framework for LLMs that supports both dynamic and static pruning.
\item We design a lightweight performance predictor based on ‌Convolutional Neural Network (CNN) to efficiently model the mapping between pruning policies and the corresponding pruned model performance using the mask matrix. This enables fast and accurate performance estimation for different pruning policies. To our knowledge, this is the first work to use pruning mask matrix for predicting performance of the pruned LLM model.
\item We conducted experiments on two representative models. The results show that with dynamic pruning ratios from 0.2 to 0.4 in Llama2-7B, PPF reduces perplexity by 33.4\%, 10.39\% and 2.59\% versus LLMPruner, LLMPruner-OWL, and LLMPruner-AlphaPruning, outperforming uniform and manually designed non-uniform methods. Under 50\% static pruning on Llama3-8B, PPF reduces perplexity by 84.78\% versus BlockPruner. Moreover, the predictor achieves second-level performance prediction, maintains prediction error below 0.0011 on unseen mask, and enables over 64$\times$ faster evaluation than test-set evaluation.
\end{itemize}
\section{Related Work}
\subsubsection{LLM Pruning.}
LLM pruning methods identify and remove redundant components to balance model efficiency and accuracy. There are several kinds of LLM pruning techniques.
Unstructured pruning removes individual weights (e.g., Wanda \cite{sun24simple}) to achieve high sparsity with minimal accuracy loss. However, its irregular zero patterns provide limited speed gains, making it less effective for LLM inference speedup. Structured pruning works at coarser levels like channels \cite{ma23llmpruner}, MHA/FFN blocks \cite{Longguang24BlockPruner}, or entire transformer blocks \cite{men2025shortgpt}. This paper focuses on channel-level structured pruning for practical speedup.
Static pruning fixes model tasks and environments throughout the process. Dynamic pruning adapts strategies to changing tasks (e.g., Probe Pruning \cite{Le25Probe}) or environments (e.g., RAP \cite{Liu25RAP}). In this work, unlike these pruning methods which are used to dynamically adapt to tasks or environments changes, PPF generates optimal policies in real-time for dynamic pruning ratios changes.
One-shot pruning removes weights in one step using handcrafted metrics (e.g., gradients \cite{ma23llmpruner}, input features \cite{An24Fluctuation}). Its success depends heavily on metric quality. Iterative pruning tests multiple policies (e.g., layer-wise sparsity \cite{sieberling2025evopress} and calibration sets \cite{Kong25Sample}) over several rounds to find optimal configurations. Due to high evaluation costs, it is typically done offline. Our performance predictor is designed to reduce this cost.
\subsubsection{Performance Prediction.}
Most existing performance predictors are designed for Neural Architecture Search (NAS) (e.g., \cite{Jawahar23AutoMoE}, \cite{Jawahar24LLMPerformance}) to discover efficient base architectures. Few address fine-tuning (e.g., \cite{ZhouL25RankAdaptor}) or cluster management (e.g., \cite{imai2024predicting}). \textbf{Notably, no prior work specifically tackles performance prediction for pruned LLMs.} Current LLM performance predictors fall into three categories: hypernetwork-based training (e.g., AutoMoE \cite{Jawahar23AutoMoE}), regression-based modeling (e.g., \cite{imai2024predicting}, \cite{ZhouL25RankAdaptor}), and LLM-based reasoning (e.g., \cite{Jawahar24LLMPerformance}). They all rely on structural parameters of full models (e.g., parameter count or layer depth) as input features. In contrast, we propose a pruning performance predictor that uses CNNs to extract features directly from pruning mask matrix and predict performance of the pruned LLMs. This predictor is novel in its use of pruning mask as input features, its CNN-based methodology, and its application to pruning scenarios, offering a new direction for LLM performance prediction.

\begin{figure*}[ht]
\centering
\includegraphics[width=0.9\textwidth]{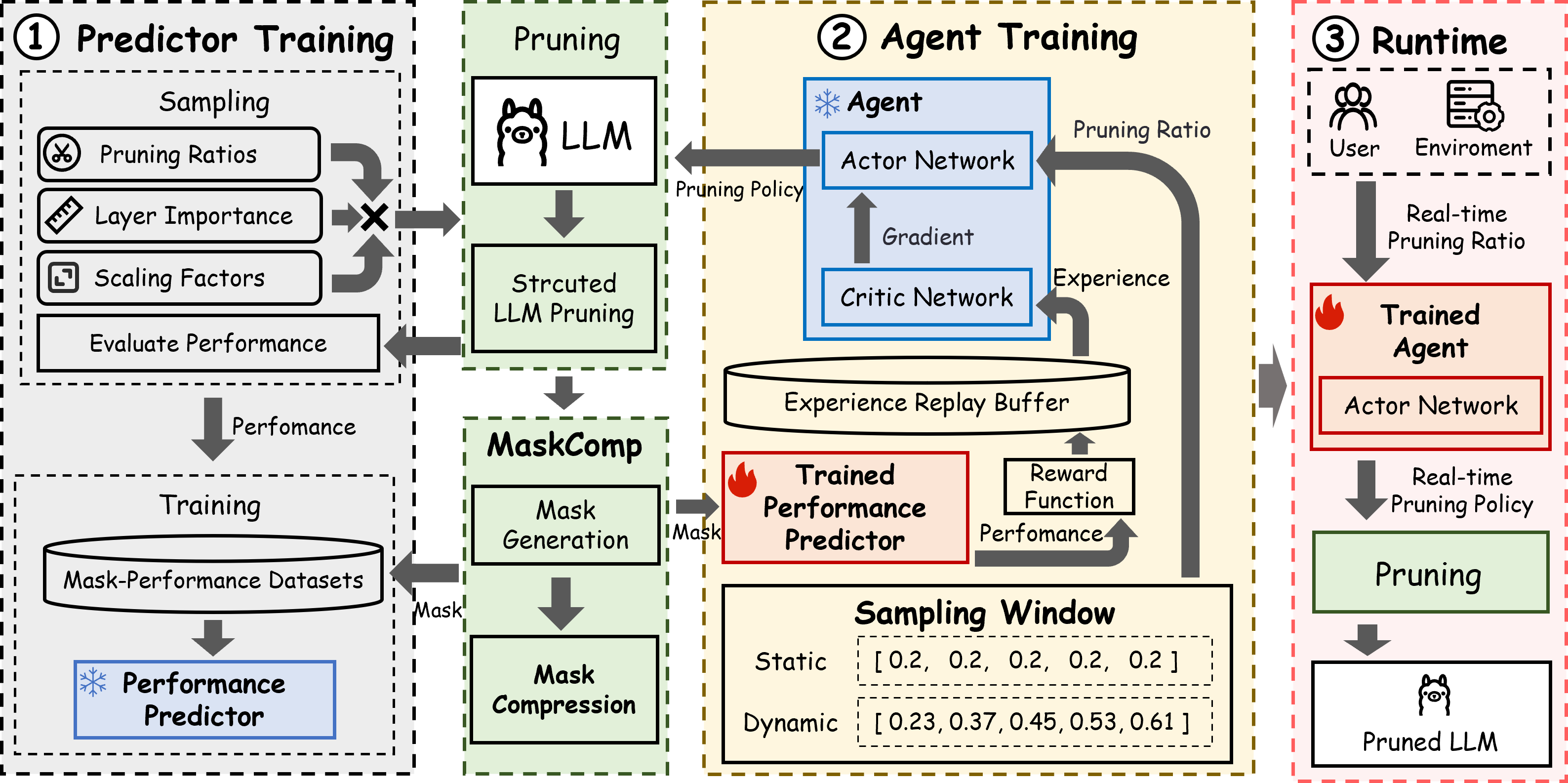} 
\caption{Overview of the PPF.}
\label{fig:ovrview}
\end{figure*}

\section{Proposed Method}
We first outline PPF's workflow, then describe its two key components in two subsections: the pruning framework and performance predictor.

As shown in Figure \ref{fig:ovrview}, we first collect some pairs of pruning mask and model performance from the models pruned using different pruning ratios, layer importance, and scaling factors. Then we train the performance predictor using these data. Based on the trained performance predictor, we train the agent used to produce pruning policies. The agent firstly takes pruning ratios one by one from a sampling window and generates pruning policies for each given ratio. Using that pruning policy, structured pruning is performed and we calculate its compressed mask. The performance predictor then quickly estimates the performance of that pruning policy based on the compressed mask. When all ratios in a sampling window are processed, the agent samples experiences from a replay buffer to update its weights. After training, the agent is deployed online to generate pruning policies in real-time.
\subsection{The Pruning Framework}
\subsubsection{The Sampling Window Strategy.}
We design a sampling window strategy to help the agent learn pruning ratio distribution patterns and generalize pruning policies across different ratios.

Suppose that the dynamic pruning range is from $\alpha$ to $\beta$, and the sampling window size is $k$. The strategy uniformly divides the dynamic pruning range $[\alpha, \beta]$ into $k$ sub-intervals $I$. Within each sub-interval $I_i$, it uniformly samples a pruning ratio $r_i$, to form a ratio sequence $R$. That process employs uniform sampling, ensuring that every pruning ratio within a sub-interval has an equal probability of being selected, thereby allowing the optimal pruning strategy for each ratio to be learned equally. The above process is shown in Eqs. \ref{eq:d_k}, \ref{eq:I_k} and \ref{eq:R_k}. 
\begin{equation}\label{eq:d_k}
\Delta=\frac{\beta-\alpha}{k}.
\end{equation}
\begin{equation}\label{eq:I_k}
I = \{ I_i = [\alpha + (i-1)\Delta, \alpha + i\Delta] \mid i=1,2,\ldots,k\}.
\end{equation}
\begin{equation}\label{eq:R_k}
R = \{ r_i \sim Uniform(I_i) \mid i=1,2,\ldots,k\}.
\end{equation}

When performing static pruning, the window parameters are set as $\alpha=\beta=S_{tar}$ (where $S_{tar}$ is a fixed pruning ratio), and the output sequence is $R=\{ r_i=S_{tar} \mid i=1,2,\ldots,k\}$. In this mode, the agent focuses on learning the optimal policy for a single pruning ratio. By repeatedly training on this fixed ratio, the agent strengthens its ability to find the optimal pruning policy for this specific setting, enabling fine-grained optimization.

An episode ends once all pruning ratios in the sampling window are processed. At this point, the agent stores the episode experience in a replay buffer. It then randomly samples a batch of experience from this buffer to update its network weights. Before starting the next episode, the pruning ratios in the sampling window are resampled. This process improves the pruning policy's generalization across the dynamic pruning range.

\subsubsection{The Agent State Space.}
Several previous works have explored Reinforcement Learning (RL) algorithms for searching model compression strategies \cite{He18AMC, Wang19HAQ}, primarily focusing on CNN architectures. In contrast, our work specifically targets LLM architectures, leveraging the Deep Deterministic Policy Gradient (DDPG) algorithm as the agent within our pruning framework. Through iterative training, this agent learns to generate real-time, optimal pruning policies. Our agent generates a pruning policy for each given pruning ratio. Its state space is defined as $S_t = r$, where $r$ represents the current given pruning ratio.

\subsubsection{The Agent Action Space.}
The effectiveness of non-uniform structured LLM pruning methods critically depends on both the layer importance computation method and the specific scaling factors.
To address this, our framework tasks the agent with choosing both the importance computation method and the ratio scaling factors, defining the pruning policy.
The agent's action space is $A_t = ( a_{i}, a_{\eta})$. $a_{i}$ is a discrete component used to represent the importance calculation method. For example, $a_{1}$, $a_{2}$, and $a_{3},$ denote the LOD from OWL \cite{yin2024outlier}, ESDs from AlphaPruning \cite{lu24alpha}, and cosine similarity from ShortGPT \cite{men2025shortgpt}, respectively. $a_{\eta}$ is a continuous component design for representing the scaling factor. For example, in this work, $a_{\eta}$ is set as values from 0.0 to 0.5. 

The layer-wise importance $H=\{ H_i \mid i=1,2,\ldots,L\}$ is determined by $a_{i}$. 
Where $L$ denotes the layer count of the model. We map the importance to layer-wise pruning ratios $S$ through Eqs \ref{eq:lamda} and \ref{eq:s}.
\begin{equation}\label{eq:lamda}
    \eta=\frac{2a_{\eta}}{Max(H)-Min(H)}.
\end{equation}
\begin{equation}\label{eq:s}
    S=\{S_i=S_{tar}+\eta(Mean(H)-H_i)\mid i=1,2,\ldots,L\}.
\end{equation}

\begin{figure}[h]
\centering
\includegraphics[width=0.45\textwidth]{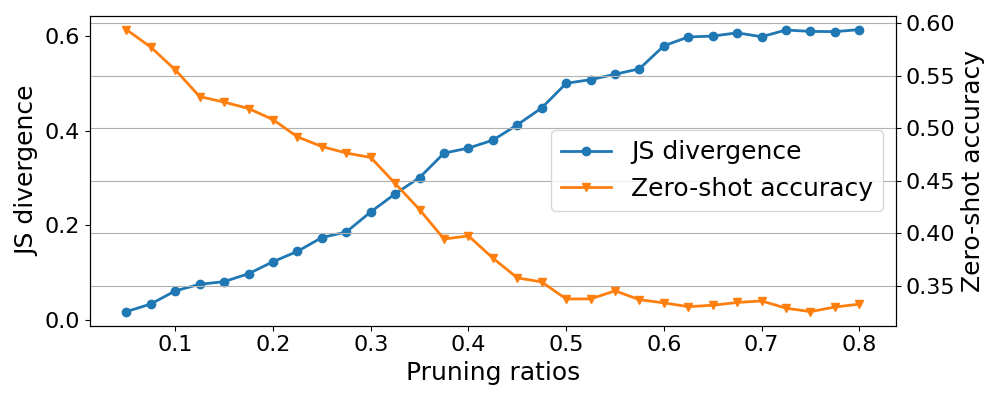} 
\caption{Comparison of JS divergence and zero-shot mean trends under varying pruning ratios.}
\label{fig:JS_zero}
\end{figure}
\subsubsection{The Agent Reward Function.}
\begin{figure*}[ht]
\centering
\includegraphics[width=0.9\textwidth]{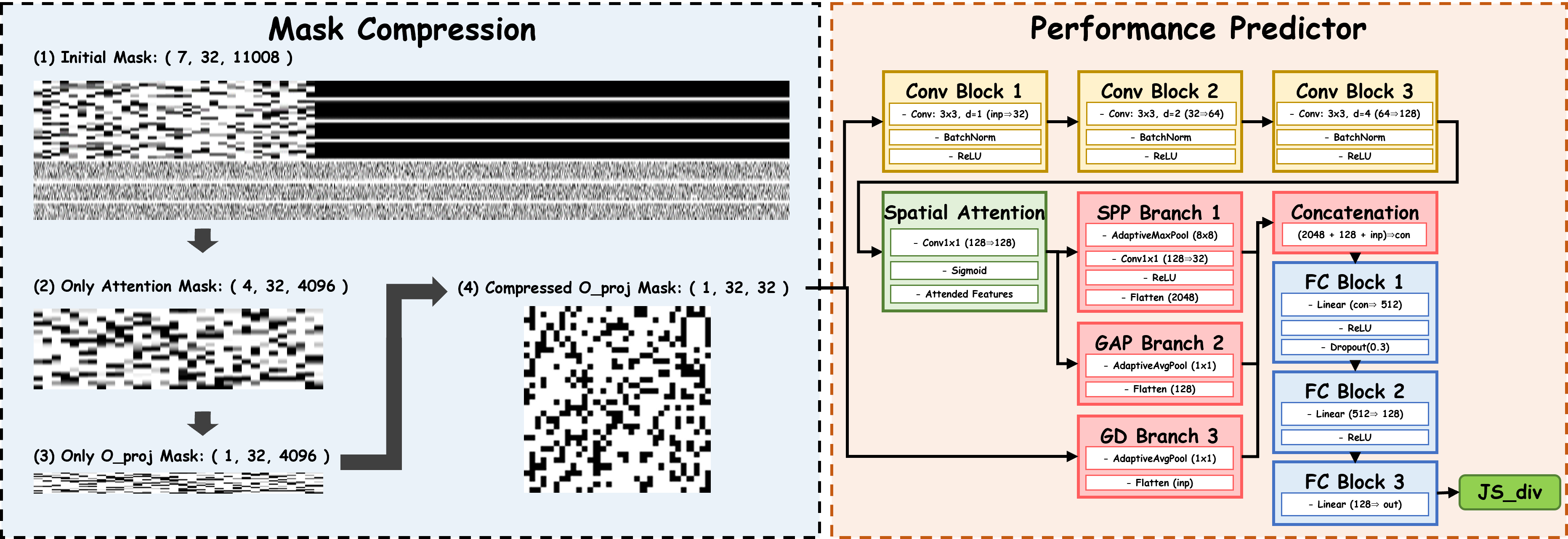} 
\caption{The pruning mask compression process and architecture of the performance predictor.}
\label{fig:PP_model}
\end{figure*}
We use the Jensen-Shannon (JS) divergence \cite{Lin91JS} to quantify the performance degradation of a pruned model, by measuring the difference in output distributions between the pruned and original models. Lower JS divergence corresponds to less performance degradation. The effectiveness of JS divergence as a comprehensive metric for evaluating compression impact is validated by \cite{Khanal24JS}. As shown in Figure \ref{fig:JS_zero}, increasing pruning ratio cause performance degradation reflected inversely by both metrics. Additionally, JS divergence is adopted because it gives a continuous value within $[0, 1]$, simplifying the training of our performance predictor.

Based on the above analysis, we propose a new comprehensive metric for evaluating pruning policies: the Performance-Parameter Ratio (PPR), as shown in Eq. \ref{eq:ppr}. 
\begin{equation}\label{eq:ppr}
  PPR=\frac{JS(P{\parallel}Q)}{r_{act}},
\end{equation}
where $P$ and $Q$ denote the original model and the pruned model, respectively, and $r_{act}$ is the actual pruning ratio.
We design $PPR$ to assess pruning policy quality independently of pruning ratio. During agent exploration, output scaling factor may yield a pruning ratio which is lower than target pruning ratio for obtaining better performance. To penalize this behavior, we introduce the pruning ratio in the denominator. As shown in Figure \ref{fig:JS_zero}, since JS divergence increases approximately linearly with pruning ratio, $PPR$ enables dimensionless comparison of pruning policies across different ratios.
Considering that lower $PPR$ values indicate superior pruning policies. Thus, we finally define the reward function as $Reward=-PPR$.

\subsection{Performance Predictor}
The performance predictor models the relationship between a pruned model (obtained using a pruning policy) and its performance. It takes the pruning mask matrix as input and outputs the predicted JS divergence. We take the mask matrix as a binary image data, where each 'pixel' represents a parameter or a parameter block. Positional information in the mask reflects the model's hierarchical structure (e.g., layer and channel indices). As shown in Figure \ref{fig:PP_model}, we design a CNN-based model that is suitable for processing such mask data to extract features and predict performance.

\subsubsection{Mask Compression}
Directly processing the mask matrix of pruned model is inefficient due to its high dimensionality. Taking Llama2-7B as an example, the initial mask shape is $(7, 32, 11008)$, where $7$ corresponds to matrix types $Q_{proj}$, $K_{proj}$, $V_{proj}$, $O_{proj}$, $Up_{proj}$, $Gate_{proj}$, $Down_{proj}$. $32$ is the layer count. $11008$ is the Feed-Forward Network (FFN) hidden dimension. Since attention dominates transformer representation capability, we first retain only attention components, thus obtaining a mask matrix with reduced dimension $(4, 32, 4096)$.

Regardless of attention implementation used Multi-Head Attention (MHA) or Group Query Attention (GQA), they all rely on $O_{proj}$ to merge attention results into the residual stream. Hence, the $O_{proj}$ matrix is the only essential output path. Therefore, we further compress the mask matrix to retain only $O_{proj}$, reducing the dimension to $(1, 32, 4096)$.

At last, considering that dependency-based pruning methods \cite{ma23llmpruner} make decisions by evaluating channel group importance (e.g., 128 dimensions per group), we further compress these groups to a final matrix with dimension $(1, 32, 32)$.
\subsubsection{CNN-based Performance Prediction Model}
The predictor needs to learn the patterns of the global pruning ratio, the layer-wise pruning distribution, and intra-layer importance from the mask data. Mask data exhibits high noise sensitivity, where minor perturbations on individual 'pixels' may cause major output variations. To address this, we use three dilated convolution layers with different dilation rates to gradually increase the receptive field while preserving spatial resolution, enabling multi-scale feature extraction.

A value of 1 in the mask represents preserved channels. Pruning critical channels can significantly impact model performance. To solve this problem, we introduce a Spatial Attention (SA) module. It learns the importance of key regions in an end-to-end manner by assigning higher weights to more important locations. The attention-weighted features are processed through both a Spatial Pyramid Pooling (SPP) module and a Global Average Pooling (GAP) layer. This captures structural patterns across different regions.
Finally, we estimate the global density of the input mask matrix using global average pooling. Features from all branches are concatenated and passed through three fully connected layers for dimensionality reduction and final output.

Notably, we found that the predictor performs best when trained with a batch size of 1. This occurs because each mask matrix represents a unique pruning strategy. With a batch size of 1, the model can quickly adapt to sample-specific features and efficiently fit the training data.
\section{Experiments}
In this section, we conduct comprehensive experiments on PPF to address three key questions:
\textbf{Q1:} Does PPF generate real-time pruning policies for dynamic pruning that outperform both uniform pruning and manually designed non-uniform policies?
\textbf{Q2:} Does PPF generate accurate pruning policies under different static pruning ratios that surpass existing static structured pruning baselines?
\textbf{Q3:} Does the performance predictor accurately evaluate model performance for unseen pruning mask while significantly reducing evaluation time?
\textbf{Q4:} How important hyperparameter in PPF affect its performance?

\subsection{Experimental Setup}
\subsubsection {Model and Dataset.} 
Our experiments are conducted using PyTorch 2.4.1 \cite{Paszke19PyTorch}, CUDA 12.1 and the models and datasets are sourced from HuggingFace \cite{Wolf19HuggingFace}. We select Llama2-7B \cite{Touvron23Llama2} and Llama3-8B \cite{Dubey24llama3} as representative models of the MHA and GQA architectures, respectively. We evaluate the zero-shot performance of the model using the following datasets: BoolQ \cite{clark2019boolq}, RTE \cite{wang2018glue}, HellaSwag \cite{zellers2019hellaswag}, WinoGrande \cite{sakaguchi2021winogrande}, ARC Easy and Challenge \cite{clark2018think}, and OpenbookQA \cite{mihaylov2018can}. The perplexity of the model is evaluated using WikiText2 \cite{merity2016pointer} and PTB \cite{Mitchell94Building}. All experiments ran on a system with 8 NVIDIA A6000 (48GB) GPUs.
\subsubsection {Baselines.} 
We select several LLM pruning baselines from officially published works (without using pre-print ones) to evaluate PPF's effectiveness: Manually designed non-uniform pruning methods include: (1)OWL \cite{yin2024outlier}, which maps the Layerwise Outlier Distribution (LOD) to a customized non-uniform sparsity ratio and (2)AlphaPruning \cite{lu24alpha} that quantifies layer importance via the shape of Empirical Spectral Densities (ESDs) in weight matrices for allocating non-uniform layerwise sparsity ratios in LLMs. The structured LLM pruning methods are as follows. (1) LLMPruner \cite{ma23llmpruner}: it constructs a structural dependency graph and uses gradient information to prune related channels. (2) ShortGPT \cite{men2025shortgpt}: it computes the Block Importance (BI) score using cosine similarity between hidden states of consecutive transformer blocks, and removes those with the lowest BI. To meet the target pruning ratio, we apply pruning from the first to the last layer of the model.  (3) BlockPruner \cite{Longguang24BlockPruner}: it measures MHA and FFN block importance using perplexity and applies a heuristic iterative strategy to progressively remove redundant these blocks. We modified its source code to stop pruning at the target ratio, as the original method struggle to reach it by adjusting pruned blocks.

\subsubsection {Implementation Details.} 
We sampled pruning ratios from 0.1 to 0.7 (step: 0.025) and scaling factors from 0.0 to 0.5 (step: 0.05). The agent is trained for 150 episodes with a sampling window size of 5. The initial action noise is 0.5, decaying at 0.95 per episode. The performance predictor is trained for 100 epochs using SGD and MSE loss.

\subsection{Dynamic Pruning Performance (Q1)}

\begin{table}[ht]
    \centering
    \begin{tabular}{c|ccc}
    \hline
         \multirow{2}{*}{Feature Support}& \multicolumn{3}{c}{Method} \\
         &   PPF& OWL& AlphaPruning\\
        \hline
         Dynamic Ratios & \ding{51} & \ding{55} (Fixed) & \ding{55} (Fixed)\\
         Real-time Control & \ding{51} & \ding{55} (Offline) & \ding{55} (Offline)\\
         Auto Importance& \ding{51} & \ding{55} (LOD) & \ding{55} (ESDs)\\
         Auto Allocation& \ding{51} & \ding{55} (Manual) & \ding{55} (Manual)\\
         Sec-response& \ding{51} & N/A &N/A \\
    \hline
    \end{tabular}
    \caption{Comparison of dynamic pruning features between PPF and non-uniform structured pruning baselines(OWL and AlphaPruning). \ding{51} indicates full support, \ding{55} indicates no support, N/A means the feature is not applicable.}
   \label{tab:Dynamic_fea}
\end{table}
Table \ref{tab:Dynamic_fea} compares key features of PPF and non-uniform pruning baseline methods (OWL and AlphaPruning). PPF supports dynamic ratio adjustment, while baselines support only fixed ratios. PPF offers real-time control, whereas baselines require offline processing. PPF automatically learns importance and allocation, but baselines rely on manual rules and allocation. PPF uniquely achieves second-level response, a capability not available in baselines.

\begin{figure}[ht]
\centering
\includegraphics[width=0.45\textwidth]{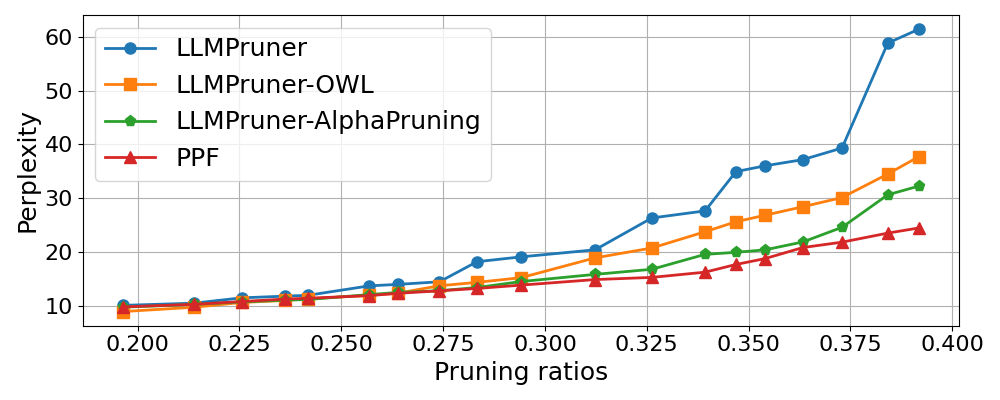} 
\caption{Comparison of perplexity on WikiText2 among PPF, LLMPruner, LLMPruner-OWL, and LLMPruner-AlphaPruning under pruning ratios (0.2–0.4).}
\label{fig:Dynamic_pruning}
\end{figure}

\begin{table*}[ht]
    \centering
    \small
    \begin{tabular}{c|c|cc|cccccccc}
         \hline
\multirow{2}{*}{\textbf{Ratio}}&  \multirow{2}{*}{\textbf{Method}}&  \multicolumn{2}{c}{\textbf{Perplexity} $\downarrow$ }&  \multicolumn{8}{|c}{\textbf{Zero-shot Tasks}  $\uparrow$ (\%)}\\
&  &  \textbf{WikiText2}&  \textbf{PTB}&  \textbf{BoolQ}&  \textbf{RTE}&  \textbf{HellaS}&  \textbf{WinoG}& \textbf{ARC-e} & \textbf{ARC-c}& \textbf{OBQA}& \textbf{Mean}\\
\hline
\multicolumn{12}{c}{\textbf{Llama2-7B}}\\
\hline
0\%  &  Dense&  5.47&  22.50&  77.73&  62.81&  57.13&  69.06&  76.30& 43.43& 31.40&59.69\\
\hline
\multirow{4}{*}{20\%} & LLMPruner&  10.18&  58.46&  62.38&  48.01&  49.01&  65.50&  \textbf{68.89}& 33.70& 28.20&50.81\\
&  ShortGPT&  227.48&  434.47&  40.40&  53.42&  28.65&  51.14&  44.75& 19.62& 15.80&36.25\\
&  BlockPruner&  11.09&  \textbf{38.31}&  62.78&  54.87&  48.60&  63.69&  64.64& 32.50& 26.80&50.55\\
&  PPF&  \textbf{9.77}&  47.15&  \textbf{69.81}&  \textbf{62.09}&  \textbf{50.62}&  \textbf{67.00}&  66.70& \textbf{35.66}& \textbf{31.80}&\textbf{54.81}\\
\hline
\multirow{4}{*}{40\%}&  LLMPruner &  64.32&  711.73&  56.23&  \textbf{54.51}&  33.63&  50.90&  41.91& 23.37& 16.40&39.57 \\
& ShortGPT &  4216.07&  2507.44&  42.70&  \textbf{54.51}&  26.90&  52.17&  31.60& 20.81& 12.60&34.47 \\
& BlockPruner & 52.77& 465.34& \textbf{61.83}& 53.42& 34.92& 53.59& 41.83& 24.23& 18.20&41.14 \\
& PPF & \textbf{35.86}& \textbf{121.21}& 51.52& 46.93& \textbf{37.96}& \textbf{58.32}& \textbf{46.80}& \textbf{28.24}& \textbf{23.40}&\textbf{41.88} \\
\hline
\multirow{4}{*}{50\%}& LLMPruner & 432.48& 974.09& 42.53& \textbf{53.79}& 27.23& 50.35& 29.92& 18.85& 13.60&33.75 \\
& ShortGPT & 17725.59& 6478.97& 37.75& 52.70& 26.30& 49.72& 26.10& 21.41& 16.80&32.97 \\
& BlockPruner & 87.05& 581.40& \textbf{60.06}& 49.45& 29.88& 52.48& 35.60& 24.14& 17.60&38.46 \\
& PPF & \textbf{54.84}& \textbf{214.11}& 51.89& 47.29& \textbf{33.52}& \textbf{55.56}& \textbf{43.01}& \textbf{25.68}& \textbf{20.60}&\textbf{39.65}\\
\hline
\multicolumn{12}{c}{\textbf{Llama3-8B}}\\
\hline
0\%  &  Dense&  6.23&  10.57&  82.07&  69.67&  60.01&  73.79&  81.52& 51.27& 33.40&64.53 \\
\hline
\multirow{4}{*}{20\%}&  LLMPruner&  15.29&  \textbf{24.43}&  \textbf{66.17}&  58.84&  45.58&  59.74&  63.34& 30.54& 18.40&48.94 \\
&  ShortGPT&  19.60&  28.79&  63.00&  55.23&  49.80&  70.40&  66.55& 39.67& 25.80&52.92 \\
&  BlockPruner&  17.77&  33.13&  66.02&  59.56&  44.99&  64.48&  64.89& 35.23& 28.20&51.91 \\
&  PPF&  \textbf{14.86}&  27.40&  62.41&  \textbf{70.75}&  \textbf{53.00}&  \textbf{71.34}&  \textbf{73.14}& \textbf{40.95}& \textbf{31.20}&\textbf{57.54} \\
\hline
\multirow{4}{*}{40\%}&  LLMPruner & 1492.33&  3576.27&  44.92&  54.51&  26.19&  50.27&  26.76& 20.39& 14.20&33.89 \\
& ShortGPT &  1152.36 &  2006.04&  45.70&  \textbf{59.20}&  \textbf{34.80}&  \textbf{60.85}&  39.75& \textbf{26.96}& \textbf{17.60}&\textbf{40.69}\\
& BlockPruner & 120.18& 257.15& 46.85& 52.70& 31.03& 50.82& 39.43& 22.26& 16.40&37.07 \\
& PPF& \textbf{47.66}& \textbf{70.00}& \textbf{57.82}& 52.70& 32.99& 54.77& \textbf{46.46}& 21.58& 16.80&40.45 \\
\hline
\multirow{4}{*}{50\%}& LLMPruner & 4223.32& 8803.03& 37.88& 51.26& 25.92& 49.96& 25.42& 20.05& 15.60&32.30 \\
& ShortGPT & 60727.25& 58210.66& 42.95& \textbf{55.23}& 28.00& 53.35& 30.95& \textbf{23.03}& 15.60&35.58\\
& BlockPruner & 984.44& 1324.00& 42.41& 52.70& 26.98& 49.48& 28.28& 21.33& 14.80&33.71 \\
& PPF& \textbf{149.78}& \textbf{291.23}& \textbf{43.30}& 52.34& \textbf{29.93}& \textbf{55.24}& \textbf{36.74}& 21.33& \textbf{16.60}&\textbf{36.50} \\
\hline
    \end{tabular}
    \caption{Comparison of perplexity and zero-shot performance across different baselines for MHA-based Llama2-7B and GQA-based Llama3-8B at pruning ratios of 20\%, 40\%, and 50\%. All models are evaluated without performance recovery strategies.}
    \label{tab:static}
\end{table*}

To evaluate the dynamic pruning performance of PPF, we set the pruning raito range from 0.2 to 0.4. Within this range, we uniformly sampled 18 pruning ratios to simulate varying dynamic pruning ratios. We compared PPF with LLMPruner combined with OWL and AlphaPruning. The mapping functions and scaling factors from their original papers were used, though they may not be optimal for all pruning ratios. As shown in Figure \ref{fig:Dynamic_pruning}, under varying pruning ratios, PPF achieves an average reduction of 33.4\% in perplexity compared to LLMPruner, and reductions of 10.39\% and 2.59\% compared to LLMPruner-OWL and LLMPruner-AlphaPruning, respectively. The result indicates that PPF achieves better dynamic pruning performance through real-time pruning policies than the uniform pruning method LLMPruner, as well as better performance than manually designed non-uniform pruning policies.

\subsection{Static Pruning Performance (Q2)}
To evaluate the static pruning performance of PPF, we consider three pruning ratios: low (20\%), medium (40\%), and high (50\%). We compare PPF with three structured LLM pruning methods of different granularities: channel-level (LLMPruner), MHA/FFN block-level (BlockPruner), and transformer block-level (ShortGPT). For fair comparison, only baselines without performance recovery strategies are included. As shown in Table \ref{tab:static}, PPF consistently reduces WikiText2 perplexity on Llama2-7B and Llama3-8B across all three pruning ratios. It also outperforms the baselines in zero-shot accuracy under most settings. For example, On the Llama3-8B model, PPF reduces WikiText2 perplexity by 84.78\% compared to BlockPruner under 50\% pruning, and improves average zero-shot accuracy by 8.73\% over ShortGPT when the pruning ratio is 20\%. Experimental results demonstrate that, under a static pruning ratio, the pruning policy found by PPF through iterative search further improves the performance of the pruned model.

\subsection{Effectiveness of the Performance Predictor (Q3)}
\subsubsection{Generalization Capability of the Predictor.}
\begin{figure}[h]
\centering
\includegraphics[width=0.45\textwidth]{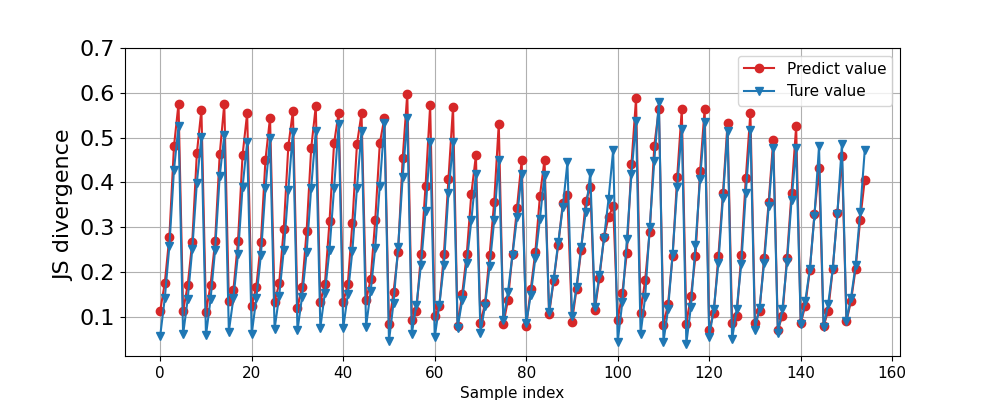} 
\caption{Prediction results of the performance predictor trained for Llama2-7B on the 20\% test set.}
\label{fig:Predictor_acc}
\end{figure}

To evaluate the predictor’s generalization to unseen pruning mask, we collected 775 samples and used 20\% of them for testing. As shown in Figure \ref{fig:Predictor_acc}, the performance predictors trained on Llama2-7B achieve promising results on the test set, with average prediction errors below 0.0011. Experimental results show that the proposed performance predictor can effectively estimate pruning performance under unseen pruning mask.

\subsubsection{Impact of the Predictor on Agent Training.}
\begin{figure}[h]
    \centering
    \begin{subfigure}[b]{0.45\textwidth}
        \includegraphics[width=\textwidth]{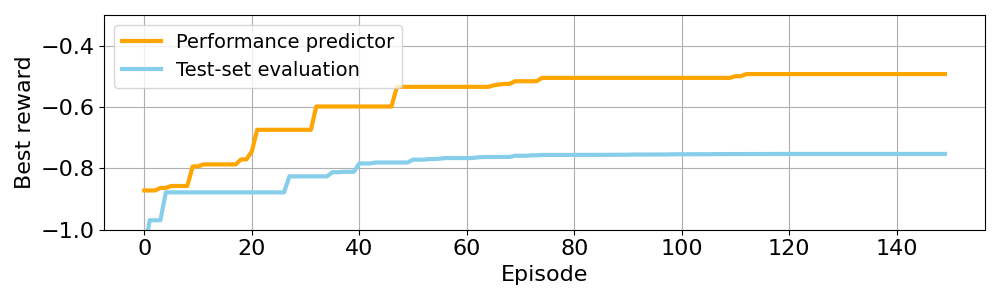}
        \caption{Comparison of best reward values.}
        \label{fig:image1}
    \end{subfigure}
    \centering
    \begin{subfigure}[b]{0.45\textwidth}
        \includegraphics[width=\textwidth]{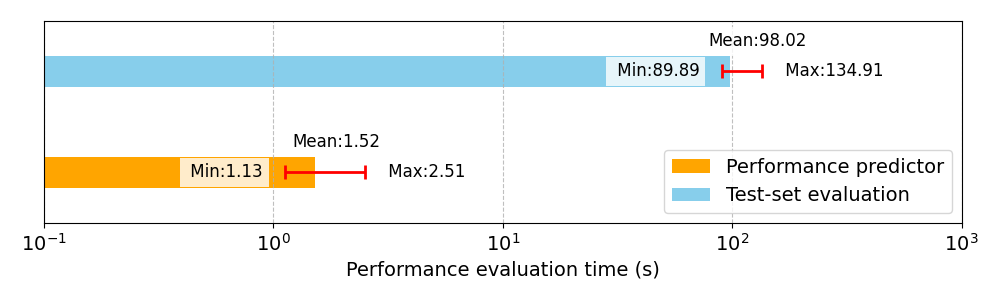}
        \caption{Comparison of evaluation time.}
        \label{fig:image2}
    \end{subfigure}

    \caption{Comparison of evaluation time and best reward per epoch: performance predictor vs. test-set evaluation during agent training.}
    \label{fig:yesnopredictor}
\end{figure}

To evaluate the effectiveness of the proposed performance predictor in accelerating the agent's training convergence and reducing evaluation time, we compare performance predictor and test-set evaluation. As shown in Figure \ref{fig:yesnopredictor}, the average performance evaluation time per episode is reduced from approximately 100 seconds to about 1 seconds, achieving a reduction from minute-level to second-level prediction and over 64$\times$ speedup in evaluation, while the best reward improves by 28.07\%. Experimental results show that the predictor significantly reduces evaluation time during agent training and accelerates convergence to the optimal solution region.

\subsubsection{Ablation Study of Mask Compression.}
\begin{figure}[h]
\centering
\includegraphics[width=0.45\textwidth]{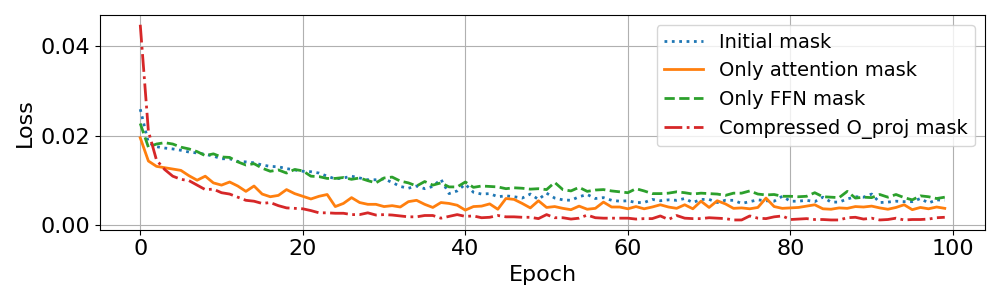} 
\caption{Comparison of loss values on the validation set across different mask compression modes.}
\label{fig:compress_loss}
\end{figure}

To verify whether the proposed mask compression strategy can reduce input complexity and improve prediction accuracy, We evaluate four compression methods: (1) initial mask, (2) mask retaining only the attention, (3) mask retaining only the FFN, and (4) mask compressed based on the $O_{proj}$ matrix. As shown in Figure \ref{fig:compress_loss}, the performance prediction achieves the best results when using the mask compressed based on the $O_{proj}$ matrix.

\begin{figure}[h]
\centering
\includegraphics[width=0.45\textwidth]{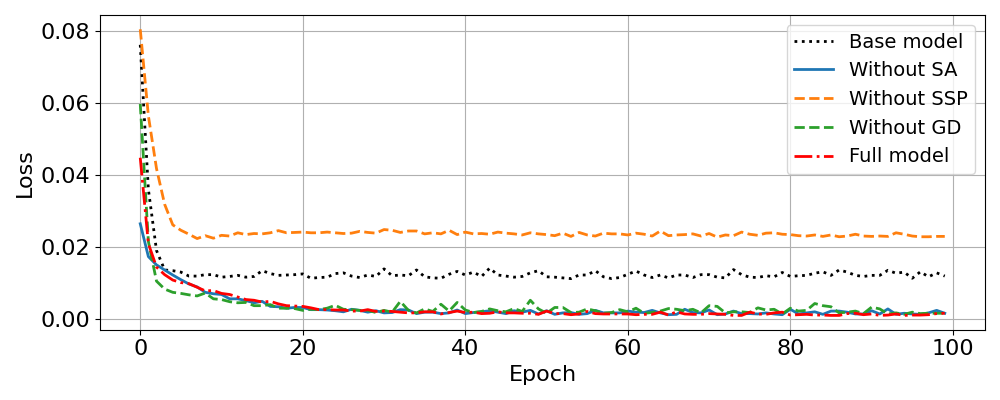} 
\caption{Comparison of loss values on the validation set after ablation of each module by the performance predictor.}
\label{fig:module_loss}
\end{figure}

\subsubsection{Ablation Study of Feature Extraction Module.}
To evaluate the impact of each module on prediction accuracy, we conducted ablation studies. Five tested variants include: (1) the base model, (2) the model without SA module, (3) the model without SPP branch, (4) the model without GD branch, and (5) the full model. As illustrated in Figure \ref{fig:module_loss}, the full model exhibits superior convergence stability in comparison with other ablation models.
\subsection{Hyperparameter sensitivity Analysis (Q4)}
\subsubsection{Impact of Action Noise and Decay in Agent.}
\begin{figure}[ht]
\centering
\includegraphics[width=0.45\textwidth]{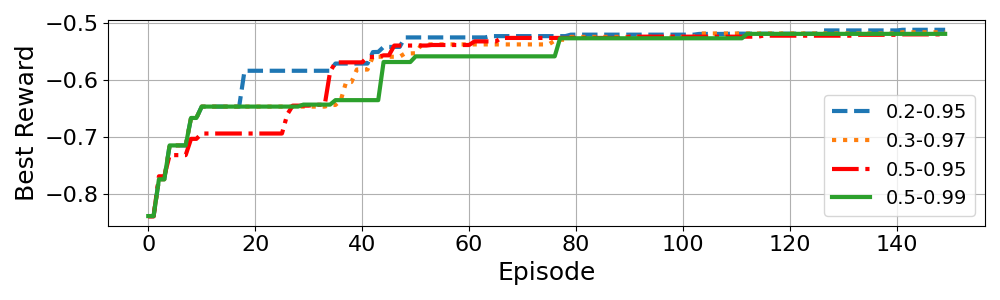} 
\caption{Effect of action noise and decay value combinations on agent best reward, where the legend entries (e.g., 0.5-0.95) denote an initial noise of 0.5 and a decay value of 0.95.}
\label{fig:noise}
\end{figure}

To evaluate the convergence stability when training PPF under varying initial conditions, we perform dynamic pruning experiments on Llama2-7B using different combinations of action noise and decay rates. As shown in Figure \ref{fig:noise}, the agent's training process is less affected by initial environmental configurations.
\subsubsection{Impact of Different sampling Window Sizes in PPF.}
\begin{figure}[ht]
\centering
\includegraphics[width=0.45\textwidth]{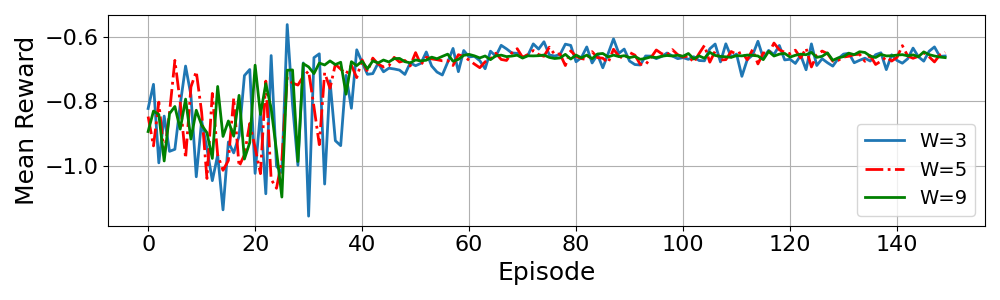} 
\caption{Effect of sampling window size on mean episode reward during agent training, where the legend entry (e.g., W=5) denotes a sampling window size of 5.}
\label{fig:window_size}
\end{figure}

We investigate the effect of sampling window size on agent training in PPF by evaluating four window settings (3, 5, and 9). Figure \ref{fig:window_size} shows that increasing the window size improves convergence stability. However, larger window size also result in higher time costs.

\section{Conclusion}
In this work, we propose PPF, a novel pruning framework for LLMs that eliminates manual design through a lightweight performance predictor and an agent for adaptive, layer-wise pruning. Experiments demonstrate its effectiveness. Since PPF's performance relies on the predictor's generalization to unseen policies, we plan to enhance it with online learning during agent training, enabling the framework to evolve dynamically and adapt to more complex pruning scenarios.

\bibliography{aaai2026}

\end{document}